\documentclass[12pt,draftclsnofoot,onecolumn]{IEEEtran}

\pdfoutput=1

\usepackage{mathrsfs}
\usepackage{cite} 
\usepackage{xspace}
\usepackage{filecontents}
\usepackage{float}
\usepackage{pgf, tikz}
\usepackage{graphicx}
\usepackage[cmex10]{amsmath}
\usepackage{mathtools} 

\usepackage{bm}
\usepackage{amssymb} 
\usepackage{array,algorithm,algorithmic}
\usepackage{amsfonts}
\usepackage{amsthm}   
\usepackage{enumerate} 
\usepackage{enumitem}
\newtheorem{definition}{Definition} 
\newtheorem{theorem}{Theorem} 
\newtheorem{lemma}{Lemma}
\newtheorem{proposition}{Proposition}
\newtheorem{remark}{Remark}

\usepackage{comment}

\usepackage{stfloats,epstopdf}
\usepackage{enumerate}
\usepackage{pgfplots}
\usepackage{caption}
\usepackage{subcaption}
\usepackage{tabularx}
\usepackage{pgfplotstable}
\usepgfplotslibrary{dateplot}
\usepackage{cite}
\usepackage{url}
\usepackage{amsbsy}
\usepgfplotslibrary{patchplots}
\usepackage{xcolor}
\usepackage{tikz}
\usetikzlibrary{shadows}
\usetikzlibrary{fadings}
\usetikzlibrary{arrows}
\usetikzlibrary{shapes,snakes}
\pgfplotsset{compat=1.7}
\usetikzlibrary{patterns}
\usetikzlibrary{spy}

\usepackage{amsmath,accents}
\DeclareMathSymbol{\widehatsym}{\mathord}{largesymbols}{"62}

\DeclareMathSymbol{\widetildesym}{\mathord}{largesymbols}{"65}

\pgfplotscreateplotcyclelist{mycyclelist_no_marks}{
solid, red, line width=1.0pt, every mark/.append style={solid, fill=red}, mark=none\\%
densely dashed, blue, line width=1.0pt, every mark/.append style={solid, fill=blue}, mark=none\\%
}

\pgfplotscreateplotcyclelist{mycyclelist}{
solid, red, line width=1.0pt, every mark/.append style={solid, fill=red}, mark=square, mark size={1}\\%
densely dashed, blue, line width=1.0pt, every mark/.append style={solid, fill=blue}, mark=triangle, mark size={1}\\%
densely dotted, line width=1.0pt, green!70!black, every mark/.append style={solid, fill=green!70!black}, mark=o, mark size={1}\\%
densely dashdotted, line width=1.0pt, brown!70!black, every mark/.append style={solid, fill=brown!70!black}, mark=diamond, mark size={1}\\%
dotted, line width=1.0pt, orange!70!black, every mark/.append style={solid, fill=orange!70!black}, mark=x, mark size={1.3}\\%
thick, solid,  red\\%
densely dashdotted, line width=1.0pt, blue\\%
densely dotted,  line width=1.2pt, green!70!black\\%
solid,  blue, every mark/.append style={solid, fill=blue}, mark=*, mark size={1}\\%
dashed, blue, every mark/.append style={solid, fill=blue}, mark=*, mark size={1}\\%
solid,  red, every mark/.append style={solid, fill=red}, mark=square*, mark size={1}\\%
dashed, red, every mark/.append style={solid, fill=red}, mark=square*, mark size={1}\\%
solid,  green!60!black, every mark/.append style={solid, fill=green!60!black}, mark=triangle*, mark size={1.2}\\%
dashed, green!60!black, every mark/.append style={solid, fill=green!60!black}, mark=triangle*, mark size={1.2}\\%
solid,  brown!70!black, every mark/.append style={solid, fill=brown!70!black}, mark=diamond*, mark size={1.2}\\%
dashed, brown!70!black, every mark/.append style={solid, fill=brown!70!black}, mark=diamond*, mark size={1.2}\\%
}
\pgfplotscreateplotcyclelist{mycyclelist2}{
solid, red, every mark/.append style={solid, fill=red}, mark=square, mark size={1}\\%
densely dashed, blue, line width=1.0pt, every mark/.append style={solid, fill=blue}, mark=triangle, mark size={1}\\%
densely dotted, line width=1.0pt, green!70!black, every mark/.append style={solid, fill=green!70!black}, mark=o, mark size={1.2}\\%
densely dashdotted, line width=1.0pt, brown!70!black, every mark/.append style={solid, fill=brown!70!black}, mark=x, mark size={1.2}\\%
thick, solid,  red\\%
densely dashdotted, line width=1.0pt, blue\\%
thick, solid,  green!70!black\\%
dashed, line width=1.0pt, brown!70!black\\
densely dashdotted, blue, every mark/.append style={solid, fill=blue}, mark=square, mark size={1}\\%
solid, blue, every mark/.append style={solid, fill=blue}, mark=triangle, mark size={1.2}\\%
densely dotted, line width=0.5pt, blue, every mark/.append style={solid, fill=blue}, mark=o, mark size={1}\\%
dashed, blue, every mark/.append style={solid, fill=blue}, mark=x, mark size={1.2}\\%
thick, dashed,  green!70!black\\%
densely dashdotted, red, every mark/.append style={solid, fill=red}, mark=square, mark size={1}\\%
solid, red, every mark/.append style={solid, fill=red}, mark=triangle, mark size={1.2}\\%
densely dotted, red, every mark/.append style={solid, fill=red}, mark=o, mark size={1}\\%
dashed, red, every mark/.append style={solid, fill=red}, mark=x, mark size={1.2}\\%
}
\pgfplotscreateplotcyclelist{mycyclelist3}{
solid,  blue, every mark/.append style={solid, fill=blue}, mark=*, mark size={1}\\%
densely dashed, red, every mark/.append style={solid, fill=red}, mark=x, mark size={1.5}\\%
}

\begin{document}
%
\title{Federated Learning Cost Disparity for IoT Devices}

\author{\IEEEauthorblockN{Sheeraz A. Alvi$^\dag$, Yi Hong$^\ddag$, Salman Durrani$^\dag$}
\IEEEauthorblockA{$^\dag$School of Engineering, The Australian National University, Canberra, ACT 2601, Australia.\\
$^\ddag$Department of Electrical and Computer Systems Engineering, Monash University, Clayton, VIC 3800, Australia.}

Emails: sheeraz.alvi@anu.edu.au, yi.hong@monash.edu, salman.durrani@anu.edu.au.}

\maketitle

\begin{abstract}
%
%
Federated learning (FL) promotes predictive model training at the Internet of things (IoT) devices by evading data collection cost in terms of energy, time, and privacy.
%
%
We model the learning gain achieved by an IoT device against its participation cost as its utility. Due to the device-heterogeneity, the local model learning cost and its quality, which can be time-varying, differs from device to device. We show that this variation results in utility unfairness because the same global model is shared among the devices.
%
%
By default, the master is unaware of the local model computation and transmission costs of the devices, thus it is unable to address the utility unfairness problem. Also, a device may exploit this lack of knowledge at the master to intentionally reduce its expenditure and thereby enhance its utility. 
%
%
We propose to control the quality of the global model shared with the devices, in each round, based on their contribution and expenditure. This is achieved by employing differential privacy to curtail global model divulgence based on the learning contribution. In addition, we devise adaptive computation and transmission policies for each device to control its expenditure in order to mitigate utility unfairness.
%
%
Our results show that the proposed scheme reduces the standard deviation of the energy cost of devices by 99\% in comparison to the benchmark scheme, while the standard deviation of the training loss of devices varies around 0.103.
\end{abstract}

\IEEEpeerreviewmaketitle


\section{Introduction}



Internet of things (IoT) is largely supported by wireless machine-type devices (MTDs) and applying machine learning techniques on the sensed data acquired by the MTDs can provide intelligent and personalized services to the user through predictive models \cite{9072101, sheeraz-2018, ALVI2014196}.
%
%
With federated learning (FL), a master device iteratively learns the global model parameters by combining the updates of local model parameters computed and then shared by the devices with the master \cite{FL-1}. From the master’s perspective, FL enables global model learning in a privacy-preserving manner, i.e., without collecting sensitive data from the MTDs. Whereas, each MTD procures a global model which is learned using disjoint data of multiple MTDs, thus a more effective model as compared to its local model. We refer to this improvement achieved in the model learning against the accumulative cost of local model computation and transmission as \textit{MTD's utility}.



In the vanilla FL setting, each MTD receives the same global model from the master, in each round, irrespective of the quality of its local model and the associated cost of computation and transmission. However, the local model quality and accumulative cost vary between MTDs due to the diversity in the training data, wireless channel, resources, etc. Furthermore, a malicious MTD may intentionally reduce the cost of computation and/or transmission, e.g., clipping training dataset size, performing fewer iterations, re-transmitting the previous update, etc. 
This malicious behavior causes utility unfairness among MTDs and damages training efficiency. Nevertheless, distinguishing this malicious behaviour from the natural channel/resource diversity is rather hard without additional information. In the existing FL settings, there are no standard of operations devised for this situation, thus the master cannot ensure pertinent operation of the MTDs.



Security and device heterogeneity are the two main issues for FL in a wireless IoT network setting. In that, the private information can still be revealed by manipulating the transmitted local models. For example, a model inversion attack analyses the differences in the shared parameters to reconstruct the training data of a device \cite{model_inversion}. In this regard, differential privacy (DP) has been shown to offer quantifiable protection against information leakage \cite{SnS-Dwork2014b}. DP is a proactive method of adding artificial noise to the data before sharing. To improve user services, a DP based personalized FL method is proposed in \cite{personal} for a wireless IoT network. DP is just employed for protection and the focus is on the adaptive training which considers device heterogeneity and data ownership. In \cite{9253545}, DP based FL algorithm is proposed for Internet of Vehicles (IoV), which minimizes the communication overhead while achieving high accuracy in a DP based secure manner. In \cite{9317806}, a DP based FL algorithm is employed to devise an incentive model based on computation, communication, and privacy cost of the devices. A higher cost yields a higher (unspecified) reward and the device's utility is the difference between the reward and the cost. Similar incentive mechanisms for FL are proposed in \cite{iot13, iot14, iot12, iot16}. From the perspective of the master's profit, some works have evaluated device contribution based on the local training delay \cite{iot13, iot14}, while other works have evaluated device contribution based on the training dataset \cite{iot12, iot16}.

\textbf{Paper Contributions:}
Prior works only focused on the expenditure or contribution of the MTDs and proposed some (typically unspecified) reward/incentive in response to the device heterogeneity. To the best of our knowledge, no prior work has jointly considered the diversity in the expenditure and contribution of the MTDs impacting the master's model learning, and addressed the unfairness among MTDs or the malicious behaviour mentioned above.
We consider a FL setting, in which multiple heterogeneous MTDs cooperate with a master. To address the utility unfairness problem among MTDs, we propose to control the quality of the global model shared with the MTDs, in each round, based on their contribution and expenditure. 
We design a utility function for MTDs to model the learning gain and cost associated with it in each round. In particular, the utility function works as a catalyst and it is used to reveal the optimal computation and transmission policies such that the learning gain versus the cost is similar for all MTDs. 
This is achieved by treating global model as a precious commodity and controlling its quality through DP. Accordingly, the master will add noise in the global model before sharing it with a MTD in proportion to the deviation of its local model from the global model. 

Our investigation leads to the following observations:

\begin{itemize}

\item The proposed scheme produces optimal computation and transmission policies for individual MTDs without any knowledge of the contribution and expenditure of other MTDs. Similarly, the master controls the global model quality without any knowledge of the cost of MTDs.

\item Our results show that the existing methodology suffers from severe utility unfairness among MTDs. The proposed scheme addresses this problem by controlling the productivity gains of MTDs and achieves similar learning gain and energy expenditure across all MTDs.

\item Our results show that the proposed scheme reduces the standard deviation of the energy cost across MTDs by 99\% in comparison to the benchmark scheme, while the standard deviation of the training loss across MTDs varies around 0.103. Also, the proposed scheme provides about 12.17\% reduction in the average energy cost of MTDs.

\end{itemize}

\section{System Model and Operation}\label{sec-sys}

\textbf{Network Setup:}
We consider a single-cell network consisting of an access point (AP) serving a set $\mathcal{K}=\{1,2,\cdot\cdot\cdot,K\}$ of MTDs located at arbitrary distances. The $k$-th MTD has a local training dataset $\mathcal{D}_k = ( \mathcal{D}_{k,1}, \mathcal{D}_{k,2}, \cdot\cdot\cdot , \mathcal{D}_{k,d_k}) \in \mathbb{R}^{d_k} = \{\mathbf{x}_{k,i} \in \mathbb{R}^s, y_{k,i} \in \mathbb{R}\}^{d_k}_{i=1}$, where $k \in \mathcal{K}$, $\mathbf{x}_{k,i}$ denote a feature vector, $y_{k,i}$ denote the corresponding label, $d_k = |\mathcal{D}_k|$ and $|\cdot|$ denote the cardinality of a set. Each MTD performs local training over its dataset and transmits the specific training parameters to the AP within a time block of $T$ secs.

\textbf{Federated Learning:}
To learn a statistical model over the datasets of all MTDs, the AP needs to find a fitting vector $\mathbf{w}_\textup{g} \in \mathbb{R}^v$ which minimizes a loss function for the given datasets. This learning task is formulated as follows \cite{sheeraz-IoT-arxiv}
\begin{equation}\label{gl-opt-prob}
\begin{aligned}
& \underset{ \mathbf{w}_\textup{g} \in \mathbb{R}^v} {\textup{minimize}}
& & \mathcal{G}(\mathbf{w}_\textup{g}) = \frac{1}{d} \sum_{k=1}^{K} d_k \mathcal{L}_k(\mathbf{w}_\textup{g}),
\end{aligned}
\end{equation}
\noindent where $d =\sum_{k=1}^{K} {d_k}$ is the total number of training samples of all MTDs, $\mathcal{G}(\cdot)$ is the empirical loss of all training samples,
\begin{equation}
\mathcal{L}_k(\mathbf{w}_\textup{g}) = \frac{1}{d_k} \sum_{i=1}^{d_k}\ell(\mathbf{w}_\textup{g},\mathbf{x}_{k,i}, y_{k,i})
\end{equation}
\noindent is the total loss function, and $\ell(\cdot)$ is a convex loss function. Therein, in the $m$-th communication round, in parallel each MTD computes the gradient of the local total loss function with respect to the global model parameters, $\nabla\mathcal{L}_k(\mathbf{w}^{(m)}_\textup{g})$, and sends it to the AP. The AP collects all gradients and computes the average as $ \nabla\mathcal{G}(\mathbf{w}^{(m)}_\textup{g}) = \frac{1}{K} \sum_{k=1}^{K} \nabla \mathcal{L}_k(\mathbf{w}^{(m)}_\textup{g})$,
and distributes it among all MTDs. Then, each MTD solves the following local loss minimization problem, and sends both the gradient $\nabla\mathcal{L}_k(\mathbf{w}^{(m)}_\textup{g})$ and the update vector $\mathbf{h}^{(m)}_k$ to the AP,
\begin{equation}\label{loc-opt-prob}
\begin{aligned}
& \underset{ \mathbf{h}^{(m)}_k \in \mathbb{R}^v} {\textup{minimize}}
& & \mathcal{F}_k(\mathbf{w}^{(m)}_\textup{g}, \mathbf{h}^{(m)}_k) = \mathcal{L}_k(\mathbf{w}^{(m)}_\textup{g} + \mathbf{h}^{(m)}_k) \\
& & & - \Big(\nabla\mathcal{L}_k(\mathbf{w}^{(m)}_\textup{g}) 
- \xi\nabla\mathcal{G}(\mathbf{w}^{(m)}_\textup{g}) \Big)^\intercal \mathbf{h}^{(m)}_k,
\end{aligned}
\end{equation}
\noindent where $(\cdot)^\intercal$ is the transpose operation, $\xi>0$ is a constant parameter \cite{sheeraz-IoT-arxiv}. The AP computes the global model as
\begin{equation}
\mathbf{w}^{(m+1)}_\textup{g} = \mathbf{w}^{(m)}_\textup{g} + \frac{1}{K} \sum_{k=1}^{K} \mathbf{h}^{(m)}_k,
\end{equation}
and broadcasts the global model towards all MTDs. After sufficient number of communication rounds, the objective function converges to a global optimal. In a given round, the computation time to compute local update by performing $j$ iterations can be given as:
\begin{equation}\label{loc-cp-time}
 T_{\text{cp},k} = j d_k \tau_k,
\end{equation}
\noindent where $\tau_k$ is the time required to process one data sample of given size \cite{sheeraz-IoT-arxiv}. Let $P_{\textup{cp},k}$ denote the power consumed by the $k$-th MTD during data processing.

\textbf{Differential Privacy:}
The AP and MTDs employ Gaussian mechanism \cite{SnS-Dwork2014b}, to achieve $(\epsilon, \delta)$-DP, by drawing a random noise vector from the Gaussian distribution such that the privacy of the model parameters is preserved.

\begin{definition}\label{def-DP}
Let $\mathcal{D}_k = ( \mathcal{D}_{k,1}, \mathcal{D}_{k,2}, \cdot\cdot\cdot , \mathcal{D}_{k,d_k}) \in \mathcal{R}$. For $\epsilon \in (0, 1)$ and $\delta > 0$, a mechanism $\mathcal{M}(\mathcal{D}_k): \mathcal{R}^{d_k} \rightarrow \mathcal{R}$, guarantees $(\epsilon, \delta)$-Differential Privacy if for all sets $\mathcal{S}$, and all parallel databases $\mathcal{D}_k$ and $\mathcal{D}_k'$ which differ by one entry, i.e., $\hbar(\mathcal{D}_k, \mathcal{D}_k') = 1$, we have
\begin{equation}
    p\{\mathcal{M}(\mathcal{D}_k') \in \mathcal{S}~|~\mathcal{D}_k'\} \leq \exp(\epsilon) p\{\mathcal{M}(\mathcal{D}_k) \in \mathcal{S}~|~\mathcal{D}_k\} + \delta,
\end{equation}
\noindent where $p\{\cdot\}$ denote the probability and $\hbar(\cdot)$ is the Hamming distance between two databases. 
\end{definition}

In Definition \ref{def-DP}, the $\epsilon$ is the privacy budget, i.e., a small value for $\epsilon$ implies more privacy and vice versa. $\delta$ is a very small probability of leaking more information than $\epsilon$.

\begin{proposition}\label{propo-privacy}
Gaussian mechanism $\mathcal{M}$ on function $f$ with sensitivity $S_f$ applied to database $\mathcal{D}_k$ achieves $(\epsilon, \delta)$-DP if 
\begin{equation}\label{sigma}
    \sigma_k   \geq \frac{1}{\epsilon} 
    \sqrt{ 2 \log \Big( \frac{1.25} {\delta} \Big) },
\end{equation}

\noindent where $\epsilon \in (0,1)$ and $\delta>0$.\label{section-DP}
\end{proposition}
\begin{IEEEproof}
The proof is given in \cite{SnS-Dwork2014b}.
\end{IEEEproof}


\vspace{0.1cm}
The variance for the noise distributions is controlled through $\sigma^2_k$ and $\sigma^2_\textup{g}$ to achieve $(\epsilon_k, \delta_k)$-DP and $(\epsilon_\textup{g}, \delta_\textup{g})$-DP at the $k$-th MTD and AP, respectively, using Proposition 1.

\textbf{Channel model:}
The AP and all the MTDs are equipped with an omnidirectional antenna. The AP allocates orthogonal radio access channel resources to MTDs for uplink transmission in a given time slot. We assume narrow-band quasi-static propagation channel between each MTD and the AP. Each channel is affected by a large-scale path loss, with path loss exponent~$\alpha$, and a small-scale Rayleigh fading, with channel coefficient $h_k$ for the $k$-th MTD. The channel gain distribution has the scale parameter $\varsigma$. We assume the channel remains unchanged over a single transmission block. The receiver antenna carries additive white Gaussian noise with zero mean and variance $\sigma^2_\text{awgn}$. Let $N_0$ denote noise spectral density. 

\textbf{Transmission:}
Each MTD transmits its local model update to the AP using the orthogonal resource blocks. Let $\mathcal{V}_v$ denote the size of the local model update in bits, where $v$ is fixed for $\mathbf{h}^{(m)}_k \in \mathbb{R}^v,~\forall\,k,m$. The transmission time for the $k$-th MTD, $T_{\textup{tx},k}$, is controlled through link transmission rate, $R_k$, i.e., 
\begin{equation}\label{t-comm}
  T_{\textup{tx},k} =  \frac{\mathcal{V}_v} {R_k}.
\end{equation}
\noindent The transmission rate, $R_k$, is given as
\begin{equation}\label{r}
R_k =  B_k \log_2 \Big(  1 + \frac {\kappa P_k |h_k|^2}{\sigma^2_\text{awgn} r_k^\alpha \Gamma} \Big),
\end{equation}
\noindent where $B_k$ is the allocated bandwidth and $P_k$ is the transmit power for the $k$-th MTD, $\kappa=\big(\frac{\text{c}}{4\pi f_c}\big)^2$ is the path loss factor, $\text{c}$ is the speed of light, $f_c$ is the center frequency, $r_k$ is the distance between MTD and the AP, and $\Gamma$ characterizes practical modulation and coding gap. Let $P_{\text{cir},k}$ denote circuit power and $\rho$ denote amplifier efficiency. The data transmission power cost $P_{\text{tx},k}$ for the $k$-th MTD can be given as \cite{tcom-sheeraz}
\begin{equation}\label{p-tx}
  P_{\text{tx},k} = \rho^{-1} P_k + P_{\text{cir},k}.
\end{equation}

\section{Proportionally-Fair Differentially Private FL}

In the proposed system, the objective of each MTD is to learn a better statistical model trained over multiple disjoint datasets of different MTDs as compared to its local model trained over much smaller dataset. The improvement achieved in statistical model learning is referred to as the learning gain. In this regard, the quality of the local model and the associated cost of computation and transmission is different for different MTDs, because of the device heterogeneity in the sensing data, wireless channel, and availability of the other resources. We quantify the utility of $k$-th MTD in $m$-th communication round, $\mathcal{U}^{(m)}_k$, by the degree of the learning gain and the accumulative energy cost associated with local model computation and transmission.

In a vanilla FL setting, each participating MTD receives the same global model update irrespective of the quality of its shared local model and the associated cost. Therein, a malicious MTD may intentionally try to reduce the cost of computation and/or transmission of the local model, which causes utility unfairness among MTDs. Consequently, even if legitimate MTDs increase their computation budget to maximize the quality of the local model to help achieve a better global model, their additional investment doesn’t pay-off productivity gains due to the poor local model contributed by the malicious MTD. Utility unfairness can also naturally be caused due to the device heterogeneity. Wherein, the local model quality and accumulative cost can be impacted by the diversity in the training data, wireless channel, computation/communication resources, etc. Nevertheless, distinguishing the malicious behaviour from the natural diversity is rather hard without additional information. Thus, the AP cannot ensure their pertinent operation. To address these challenges, we propose to preserve global model using DP. We design policies for the AP and the MTDs, where the AP strives to ensure utility fairness among MTDs and the MTDs try to maximize their utility. In particular, the utility function works as a catalyst and it is used to reveal the optimal computation and transmission policies, such that the learning gain versus the cost is similar for all devices.

\subsection{MTD's Local Training Convergence and Cost}

In each communication round, all MTDs perform multiple iterations to solve the problem in \eqref{loc-opt-prob} with an accuracy of $\Phi$. 

\begin{definition}\label{def-accuracy}
For the $k$-th MTD, in the $m$-th communication round, we define the accuracy $\Phi \in (0,1)$ of the solution $\mathbf{h}^{(m),(j)}_k$ to the local problem in \eqref{loc-opt-prob} after $j$ iterations as
\begin{equation}\label{loc-accuracy}
    \Phi \geq \frac{\mathcal{F}_k(\mathbf{w}^{(m)}_\textup{g},\mathbf{h}^{(m),(j)}_k)-\mathcal{F}_k(\mathbf{w}^{(m)}_\textup{g},\mathbf{h}^{*(m)}_k)  } {\mathcal{F}_k(\mathbf{w}^{(m)}_\textup{g},\mathbf{0})-\mathcal{F}_k(\mathbf{w}^{(m)}_\textup{g},\mathbf{h}^{*(m)}_k)},
\end{equation}
\noindent where $\mathbf{h}^{*(m)}_k$ is the optimal solution of the problem in \eqref{loc-opt-prob}.  
\end{definition}

We obtain a lower bound on the number of iterations, $j_\textup{min}$, required to solve local problem in \eqref{loc-opt-prob} with an accuracy $\Phi$.

\begin{theorem}\label{theorem-iteration}
If the objective function $\mathcal{F}_k(\cdot)$ in \eqref{loc-opt-prob} is twice-continuously differentiable $\mu$-strongly convex and its gradient $\nabla \mathcal{F}_k(\cdot)$ is $L$-Lipschitz continuous, then $k$-th MTD employing the gradient method with a step size $\eta > \frac{L}{2}$ needs to perform 
\begin{equation}\label{iter-bound}
  j_{\textup{min},k} \geq \frac{ \log(\Phi) } 
            { \log \big( 0.5 \eta^2 L^2 - \eta L  + 1 \big) }
\end{equation}
\noindent iterations to solve the problem in \eqref{loc-opt-prob} with an accuracy $\Phi$.
\end{theorem}
\begin{IEEEproof}
The proof is provided in \cite{sheeraz-IoT-arxiv}.
\end{IEEEproof}


\vspace{0.1cm}
Finally, from \eqref{loc-cp-time}, \eqref{t-comm}, and \eqref{p-tx}, the total energy cost of the $k$-th MTD in the $m$-th communication round can be given as
\begin{equation}\label{energy-cost}
  E^{(m)}_{\text{cp+tx},k} = j_k d_k \tau_k P_{\text{cp},k} + 
  \frac{\mathcal{V}_v (\rho^{-1} P^{(m)}_k + P_{\text{cir},k}) } 
  {B_k \log_2 \Big(  1 + \kappa \frac {P^{(m)}_k |h^{(m)}_k|^2}{\sigma^2_\text{awgn} r_k^\alpha\Gamma}\Big)},
\end{equation}
\noindent where $x^{(m)}$ indicates that $x$ varies from one round to the next.

\subsection{Utility Fairness Policy for AP}

After collecting local updates from the MTDs, the AP computes the true global model $\mathbf{w}^{(m)}_\textup{g}$. The quality of the local model update differs across MTDs. Sharing the same global model with all MTDs results in utility unfairness. In this setting, we propose to control the quality of the global model shared with different MTDs, in each communication round, based on their contribution towards the global model computation. This is achieved by employing DP to curtail global model divulgence based on the learning contribution. Therein, the AP adds noise in the global model before sharing it with a given MTD in proportion to the deviation of its local model from the global model. The AP relies only on the local model quality to decide the level of noise to be added. It is because in practical settings the computation and transmission cost of MTDs cannot not be quantified effectively at the AP.

The system guarantees at least $(\epsilon_\text{g},\delta_\text{g})$-DP. The noise variance is increased further in proportion to the quality of the individual MTD's local model update. Accordingly, in the $m$-th communication round, the noise vector for the $k$-th MTD is drawn from distribution $\mathcal{N}(0, S^{2,(m)}_{f_\textup{g}}\widehat{\sigma}^{2,(m)}_{\textup{g},k})$, where 
\begin{equation}
    \widehat{\sigma}^{(m)}_{\textup{g},k} \geq \frac{1}{\epsilon_\textup{g} ( 1 - \mathcal{E}^{(m)}_k \theta) }  \sqrt{2\log \Big( \frac{1.25 }{\delta_\textup{g} } \Big) },
\end{equation}
\noindent and $\mathcal{E}^{(m)}_k \in [0,1]$ captures how different the local model is from the true global model, and $\theta \in [0,1]$ calibrates the impact of $\mathcal{E}^{(m)}_k$ on $\epsilon_\textup{g}$. We refer to $\mathcal{E}^{(m)}_k$ as the deviation factor and employ the Cosine Similarity measure to quantify it as follows:
\begin{equation}\label{dev-factor}
    \mathcal{E}^{(m)}_k = 1 - 
    \frac{ \textup{sim}\big(\mathbf{w}^{(m)}_\textup{g},\mathbf{h}^{(m)}_k\big)}
    {\underset{ \forall \, k \in \mathcal{K}}  {\max} \big\{\text{sim}\big(\mathbf{w}^{(m)}_\textup{g},~\mathbf{h}^{(m)}_k \big) \big\} },
\end{equation}
\noindent where $\textup{sim}(\cdot)$ is the Cosine Similarity operation. In the $m$-th round the $k$-th MTD receives the deviation factor $\mathcal{E}^{(m)}_k$ and the global model update $\mathbf{w}^{(m)}_{\textup{g},k} = \mathbf{w}^{(m)}_\textup{g} + \mathbf{n}^{(m)}_k$, where $\mathbf{n}^{(m)}_k \sim \mathcal{N}(0, S^{2,(m)}_{f_\textup{g}}\widehat{\sigma}^{2,(m)}_{\textup{g},k})$ is the noise vector, from the AP.
 
\subsection{Utility Maximization Policy for MTDs}

The utility of a MTD decreases when the AP adds more noise to its global model update, which is directly proportional to the MTD's deviation factor. Hence, the learning gain is proportional to the deviation factor, i.e., local model's quality. Using data fitting analysis, we model the deviation factor with the MTD energy cost as follows
\begin{equation}\label{mod}
\mathcal{E}^{(m)}_{\textup{mod}, k} = \beta^{(m)}_{1,k} \exp \Big( - \frac{1}{\beta^{(m)}_{2,k}} T^{(m)}_{\text{cp},k} P^{(m)}_{\text{cp},k} \Big),
\end{equation}
\noindent where $\beta^{(m)}_{1,k}, \beta^{(m)}_{2,k} > 0$ are model parameters. The values for $\beta^{(m)}_{1,k}$, $\beta^{(m)}_{2,k}$ are estimated using mean-square-error method:
\begin{equation}
 \hat{\beta}^{(m)}_{1,k}, \hat{\beta}^{(m)}_{2,k} \leftarrow \underset{ \beta^{(m)}_{1,k}, \beta^{(m)}_{2,k}} {\textup{argmin}}
|\mathcal{E}^{(m)}_k - \mathcal{E}^{(m)}_{\textup{mod}, k} |^2.
\end{equation}
These values of $\hat{\beta}^{(m)}_{1,k}$, $\hat{\beta}^{(m)}_{2,k}$ are used in the next $(m+1)$-th round to maximize its expected-utility. This strategy closely resembles the risk-aversion in expected-utility theory \cite{exp-utility}, wherein the utility function is modelled as concave in cost. We model the utility function as the following concave function:
\begin{equation}\label{MTD-utility}
\mathcal{U}_k = - \mathcal{E}_{\textup{mod}, k} + \beta_{1,k} - E_{\textup{cp+tx}, k} (E_{\textup{cp+tx}, k} - \varrho),
\end{equation}
\noindent where parameter $\varrho > 0$ captures the relationship between the utility and MTD's energy cost. In \eqref{MTD-utility}, the first two terms jointly represent the relative quality of the local model and the other term represents the impact of the total energy cost.
Using $\hat{\beta}^{(m)}_{1,k}$, $\hat{\beta}^{(m)}_{2,k}$, MTD solves the following problem to obtain the optimal computation and transmission policies which will yield the maximum utility in the next $(m+1)$-th round,
\begin{subequations}\label{utility-max-opt-prob-tmp}
\begin{alignat}{2}
& \hspace{-1.0cm}\underset{ \substack{\Phi^{(m+1)}_k,~j^{(m+1)}_k, \\ P^{(m+1)}_k,~R^{(m+1)}_k} }   {\textup{maximize}}
& &  \mathcal{U}_k\big(\hat{\beta}^{(m)}_{1,k}, \hat{\beta}^{(m)}_{2,k}, j^{\S,(m+1)}_k, P^{(m+1)}_k \big)         \label{utility-max-opt-prob-tmp-a}                  \\
& \hspace{-0.6cm}\textup{subject to}
& &    T^{(m+1)}_{\textup{cp}, k} + T^{(m+1)}_{\textup{tx}, k} \leq T, \label{utility-max-opt-prob-tmp-b} \\
& & &  P_{\textup{min}, k} \leq P^{(m+1)}_k \leq P_{\textup{max}, k}, \label{utility-max-opt-prob-tmp-d}\\
& & &  j_{\textup{min}, k} \leq j^{(m+1)}_k \leq j_{\textup{max}, k}, \label{utility-max-opt-prob-tmp-f}\\
& & &  0 \leq \Phi^{(m+1)}_k \leq 1, \label{utility-max-opt-prob-tmp-e}\\
& & &  0 \leq R^{(m+1)}_k. \label{utility-max-opt-prob-tmp-g}%
\end{alignat}
\end{subequations}
\noindent where \eqref{utility-max-opt-prob-tmp-b} states that accumulative computation and transmission time should not exceed the delay bound. The remaining constraints reflect practical range of values for the design variables, where $P_{\textup{min}, k}$ is the transmit power of the $k$-th MTD required to perform $j_{\textup{min}, k}$ iterations, i.e., 
\begin{equation}\label{P-min}
P_{\textup{min}, k} =  \frac {\sigma^2_\textup{awgn} r^\alpha_k \Gamma } {\kappa |h_k|^2}  \bigg( \exp \Big( \frac{\mathcal{V}_v \log(2)} 
  {B_k (T - j_{\textup{min}, k} d_k \tau)} \Big) -  1 \bigg),
\end{equation}
\noindent and $j_{\textup{max}, k}$ is the upper bound on the number of iterations the $k$-th MTD can perform, i.e., 
\begin{equation}\label{j-max}
j_{\textup{max}, k} = \frac{1}{d_k \tau_k} \Big( T  {-}  \frac{\mathcal{V}_v \log(2) } 
  {B_k \log \big(  1 {+} \kappa \frac {P_{\textup{max}, k} |h_k|^2}{\sigma^2_\text{awgn} r_k^\alpha\Gamma}\big)} \Big).
\end{equation}

\begin{remark}\label{remark-1}
Using Theorem 1, we can compute the bound on the accuracy with which the local problem will be solved after a given number of iterations. Similarly, the link transmission rate can be computed through the transmit power.
\end{remark}

Based on Remark 1, substituting $T^{(m+1)}_{\textup{cp}, k}$, $T^{(m+1)}_{\textup{tx}, k}$, and $P^{(m+1)}_{\textup{tx}, k}$ from \eqref{loc-cp-time}, \eqref{t-comm}, and \eqref{p-tx}, for an arbitrary communication round, the problem \eqref{utility-max-opt-prob-tmp} can equivalently be given as:
\begin{subequations}\label{utility-max-opt-prob}
\begin{alignat}{2}
& \underset{ j_k,~P_k }  {\textup{maximize}}
& & \hspace{0.3cm} \mathcal{U}_k\big(\hat{\beta}_{1,k}, \hat{\beta}_{2,k}, j_k, P_k \big)                         \label{utility-max-opt-prob-a}  \\
& \textup{subject to}
& & \hspace{0.3cm}   j_k d_k \tau_k + 
  \frac{\mathcal{V}_v \log(2) } 
  {B_k \log \big(  1 {+} \kappa \frac {P_k |h_k|^2}{\sigma^2_\text{awgn} r_k^\alpha\Gamma}\big)} \leq T, \label{utility-max-opt-prob-b}\\
& & & \hspace{0.3cm}  P_{\textup{min}, k} \leq P_k \leq P_{\textup{max}, k}, \label{utility-max-opt-prob-d}\\
& & & \hspace{0.3cm} j_{\textup{min}, k} \leq j_k \leq j_{\textup{max}, k}.\label{utility-max-opt-prob-e}
\end{alignat}
\end{subequations}
The problem in \eqref{utility-max-opt-prob} is a non-convex optimization problem.

\begin{lemma}\label{lemma-convex-transform}
The optimization problem in \eqref{utility-max-opt-prob} can be transformed into an equivalent convex problem. Thus, a globally optimal solution
exists for the problem in \eqref{utility-max-opt-prob}.
\end{lemma}

\begin{IEEEproof}
The proof is provided in \cite{sheeraz-IoT-arxiv}.
\end{IEEEproof} 

From Lemma \ref{lemma-convex-transform}, we have the following equivalent convex problem for \eqref{utility-max-opt-prob}:
\begin{subequations}\label{utility-max-opt-prob-convex}
\begin{alignat}{2}
& \underset{ j_k,~Z_k }  {\textup{minimize}}
& & \hspace{0.3cm} -\mathcal{U}_k\big(\hat{\beta}_{1,k}, \hat{\beta}_{2,k}, j_k, Z_k \big)                         \label{utility-max-opt-prob-a}  \\
& \textup{subject to}
& & \hspace{0.3cm}   j_k d_k \tau_k + \frac{\mathcal{V}_v \log(2)}  {B_k Z_k} \leq T, \label{utility-max-opt-prob-b}\\
& & & \hspace{0.3cm}  Z_{\textup{min},k} \leqslant Z_k \leqslant Z_{\textup{max},k}, \label{utility-max-opt-prob-d}\\
& & & \hspace{0.3cm}  j_{\textup{min}, k} \leq j_k \leq j_{\textup{max}, k}.\label{utility-max-opt-prob-e}
\end{alignat}
\end{subequations}
\noindent where $Z_k = \log \big(  1 +  { \kappa P_k |h_k|^2}
{(\sigma^2_\textup{awgn} r^\alpha_k \Gamma)^{-1} } \big)$, $Z_{x,k} = \log \big(  1 +   {\kappa P_{x,k} |h_k|^2}{(\sigma^2_\textup{awgn} r^\alpha_k \Gamma )^{-1}} \big)$, $x \in \{\text{min, max\}}$. \vspace{0.1cm} 

\begin{remark}\label{remark-2}
The solution to the problem in \eqref{utility-max-opt-prob-convex}, i.e., the optimal values of $j_k$ and $Z_k$, yield the optimal solution to the problem in \eqref{utility-max-opt-prob} which will maximize its objective function.
\end{remark}

Based on Remarks \ref{remark-1} and \ref{remark-2}, and Lemma \ref{lemma-convex-transform}, the solution to the problem in \eqref{utility-max-opt-prob-convex} yields the solution to the problem in \eqref{utility-max-opt-prob-tmp} as given by the following theorem.
\begin{theorem}
In solving the optimization problem \eqref{utility-max-opt-prob-convex}, the optimal transmission rate, $R^*_k$, is given by
\begin{equation}\label{optimal-R}
R^*_k = B_k \log_2 \big(  1 {+} \kappa P^*_k |h_k|^2 \sigma^{-2}_\textup{awgn} r^{-\alpha}_k \Gamma^{-1}  \big)
\end{equation}
\noindent where the optimal transmit power, $P^*_k$, is given by
\begin{equation}\label{optimal-P}
P^*_k = \sigma^2_\textup{awgn} r^\alpha_k \Gamma \kappa^{-1}
|h_k|^{-2}  \big( \exp(Z^*_k) -  1 \big),
\end{equation}
\noindent where
\begin{equation}\label{optimal-z}
Z^*_k=
    \begin{cases}
      \max \big(Z_{\textup{min}, k},~\hat{Z}_k \big),    
      & \textup{if} \; \hat{Z}_k < Z_{\textup{max}, k}, \\
      Z_{\textup{max}, k},      &   \textup{otherwise,}
    \end{cases}
\end{equation}
\noindent where $\hat{Z}_k$ is given by numerically solving following equality 
\begin{multline}\label{optimal-z-hat}
\hspace{-0.2cm}\Big( 2 P_{\textup{cp},k} \big( T {-} \mathcal{V}_v \log(2) B^{-1}_k \hat{Z}^{-1}_k  \big)
{+}  2 \mathcal{V}_v b_k \hat{Z}^{-1}_k  \big( \exp  ( \hat{Z}_k ) {+} c_k  \big) \\ {-} \varrho  \Big) 
\Big( \frac {B_k b_k}{P_{\textup{cp},k} \log(2)} \big( {(\hat{Z}_k - 1) \exp( \hat{Z}_k ) - c_k} \big) + 1 \Big)
\\
= \beta_{1,k} \beta^{-1}_{2,k} \exp \Big( P_{\textup{cp},k} \beta^{-1}_{2,k}
  \big( \mathcal{V}_v  B^{-1}_k \hat{Z}^{-1}_k \log(2) - T \big) \Big),
\end{multline}
\noindent where $b_k = \frac{ \sigma^2_\textup{awgn} r^{\alpha}_k \Gamma \log(2)}{\rho \kappa B_k |h_k|^2}, c_k = \frac{\rho \kappa |h_k|^2 P_{\textup{cir},k} } {  \sigma^2_\textup{awgn} r^{\alpha}_k \Gamma} - 1$,
and the optimal number of iterations, $j^*_k$, to perform is given by 
\begin{equation}\label{optimal-j}
j^*_k = \min \Big( d^{-1}_k \tau^{-1}_k \big( T {-} \mathcal{V}_v  B^{-1}_k Z^{*-1}_k \log(2) \big),~j_{\textup{max}, k}\Big),
\end{equation}
\noindent and the optimal accuracy, $\Phi^*_k$, is bounded by
\begin{equation}\label{optimal-phi}
 \Phi^*_k \leq \exp \Big( \big( j^*_k + 1 \big)  \log \big( 0.5 \mu \eta^2 L - \mu \eta  + 1 \big) \Big)
\end{equation}
\end{theorem}
\begin{IEEEproof}
The proof is provided in \cite{sheeraz-IoT-arxiv}.
\end{IEEEproof}

\section{Simulation Results}\label{sec-results}

In this section, we first present the learning performance of the proposed scheme, and then study the impact of device heterogeneity on the energy expenditure. To model the variable computation cost of local training at the MTDs, we keep the dataset size same at each MTD and the quality of the local model is controlled through the number of iterations, which is inline with prior works \cite{personal, 9253545, 9317806}. For simulations we consider the MNIST dataset available for digit recognition task and a neural network for training with an input layer with 784 units, two hidden layers (the first with 128 units, the second with 64 units) each using ReLu activation, then an output layer with 10 units, and the softmax output. The total number of parameters is 109,375, each represented by one byte, i.e., $\mathcal{V}_v$ = 875 kbits. The batch size is set to 128 for all MTDs. The sensitivity of the data varies around 0.01. Unless specified otherwise, the parameters values shown in Table \ref{para-table} are adopted.

\begin{table}[]
\centering
\vspace{0.1cm}\caption{System parameter values.}
\label{para-table}
\begin{tabular}{|c|c||c|c||c|c|} \hline
\textbf{Sym.} & \textbf{Value}  & \textbf{Sym.}  & \textbf{Value}  & \textbf{Sym.}  & \textbf{Value} \\ \hline
 $\epsilon_\text{g}$ & 0.95     & $P_{\textup{cir},k}$ & 82.5 mW, $\forall k$    & $f_c$    & 32 MHz\\
 $\theta$         & 0.6   & $P_{\textup{max},k}$ & 0 dB, $\forall k$  & $\Gamma$ & 9.8 dB \\
 $K$                 & 10       & $T$                & 0.75 ms  & $\delta_\text{g}$  & $10^{-5}$   \\
 $\alpha$            & 4        & $r_k$              & \{50-200\} m, $\forall k$ & $\tau$   & 7.5 ns/b \\
 $\varrho$           & 0.5      & $\mathcal{V}_v$    & 875 kbits & $B$                & 250 KHz     \\
 $\varsigma$         & 1        & $P_{\textup{cp},k}$& 96 mW, $\forall k$ & $j_{\textup{min},k}$    & 10, $\forall k$ \\
 $\rho$              & 0.45     & $N_0$              & $-$174 dBm      & $S_f$ & $\approx$ 0.01 \\ \hline
%
%
\end{tabular}
\vspace{-0.25cm}
\end{table}

Recent works \cite{personal, 9253545, 9317806} are the most relevant to our proposed scheme. Although the objectives are different, the system models specifying the underlying FL and DP implementation are similar to our considered system. In this regard, our objective is to identify the utility unfairness issue among MTDs and these models suffice to demonstrate it. Once unfairness issue is divulged, we analyse the performance of the proposed scheme to counter that. When our considered system is applied, the design for the DP based FL models in \cite{personal, 9253545, 9317806} can equivalently be represented by the following benchmark scheme.

\textbf{\textit{Benchmark scheme}}:
For the benchmark scheme, the algorithm aims to achieve a fixed $(\epsilon_\text{g},\delta_\text{g})$-DP for both local and global models sharing. This is the minimum level of DP the proposed scheme already guarantees. In the benchmark scheme, each MTD tries to perform the maximum number of iterations ($\leq j_\textup{max}$) possible and transmit the noisy local model to the AP under the given channel realization and delay constraint. The AP receives the local models and generates the global model. A noisy version of this global model, which is same for each MTD, providing $(\epsilon_\text{g},\delta_\text{g})$-DP is then sent to all MTDs. The strategy followed by the benchmark scheme implements a generic differentially private FL algorithm with computation and transmission cost control. This strategy is essentially the same as in the state-of-the-art in \cite{personal, 9253545, 9317806}. The corresponding optimization problem for the benchmark scheme is omitted here for brevity. For fair comparison, the dataset of a given MTD and channel realizations in a given round are kept the same for both schemes.

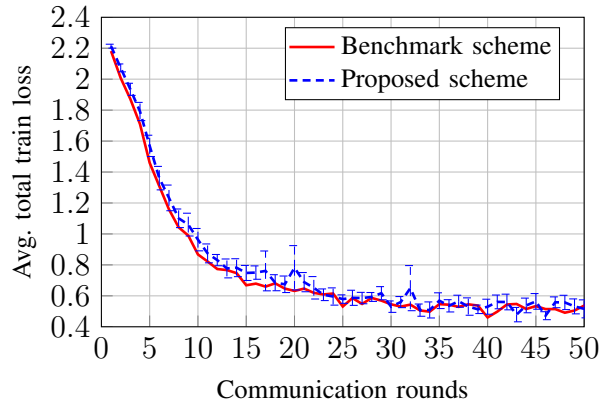
\begin{figure}[t]
\centering
    \begin{tikzpicture}[spy using outlines={ chamfered rectangle, magnification=2.0, width=1.25cm, height=2.0cm,, connect spies}]
    \begin{axis}[
     each nth point=1,
     height=5.7cm, width=8.0cm,
     legend cell align=left,
     legend style={inner xsep=1pt, inner ysep=1pt,at={(0.95,0.85)},anchor=east,font=\small, legend columns=1, draw, fill},
     cycle list name = mycyclelist_no_marks,
     mark repeat={1},
     grid=both,
     label style={font=\small},
     xtick pos=left,
     ytick pos=left,
     xmin=0, xmax=50, xtick={0,5,10,15,20,25,30,35,40,45,50},
     xlabel= Communication rounds,
     ylabel= Avg. total train loss,
     ylabel style={yshift=-0.5ex},
     ymin=0.4, ymax=2.4, ytick={0.4,0.6,0.8,1.0,1.2,1.4,1.6,1.8,2.0,2.2,2.4},
     ticklabel style={
        /pgf/number format/fixed,
        /pgf/number format/precision=5
     }
     ]
             \addplot+ [
                error bars/.cd,
                    y explicit,
                    y dir=both,
            ] table [
                x=Round,
                y=NUM,
            ] {data_b.txt};
            
            \addplot+ [
                error bars/.cd,
                    y explicit,
                    y dir=both,
            ] table [
                x=Round,
                y=NUM,
                y error plus expr=\thisrow{CI-H}-\thisrow{NUM},
                y error minus expr=\thisrow{NUM}-\thisrow{CI-L},
            ] {data_p.txt};
    \legend{Benchmark scheme, Proposed scheme}
    \end{axis}
        \end{tikzpicture}
\caption{Average MTD total train loss, $\frac{1}{K}\sum_{i=1}^{K} \mathcal{L}^{(m)}_k$, over communication rounds for the benchmark and proposed schemes.}
\label{avg_loss}
\vspace{-0.25cm}
\end{figure}

We first perform the comparative convergence analysis of the proposed scheme with the benchmark scheme. Fig.~\ref{avg_loss} plots the average MTD total train loss, $\frac{1}{K}\sum_{i=1}^{K} \mathcal{L}^{(m)}_k$, over the communication rounds for the benchmark and proposed schemes. The simulations were run for 200 rounds but only first 50 are shown in Fig.~\ref{avg_loss} for better clarity. Although the channel heterogeneity (including the path loss) exists among MTDs, in any given round, the average loss and the standard deviation in loss across different MTDs are very similar for both schemes. Specifically, the average training loss is only 6.26\% higher for the proposed scheme. This shows that the overall learning performance does not suffer from the proposed MTD-wise adaptive global model quality control. In addition, a small standard deviation, i.e., around 0.103, in training loss indicates that the learning experience is fair among MTDs.

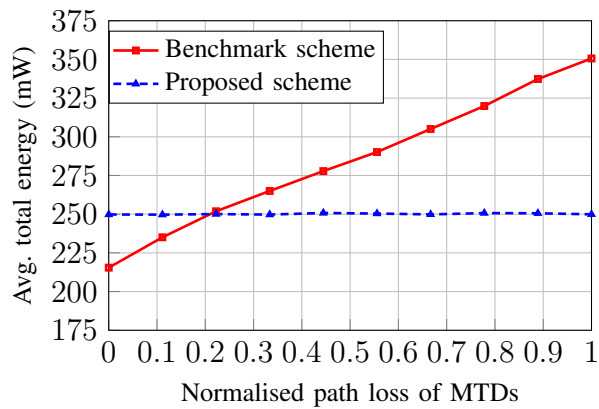
\begin{figure}[t]
\centering
    \begin{tikzpicture}[spy using outlines={ chamfered rectangle, magnification=2.0, width=1.25cm, height=2.0cm,, connect spies}]
    \begin{axis}[
     height=5.7cm, width=8.0cm,
     legend cell align=left,
     legend style={inner xsep=1pt, inner ysep=1pt,at={(0.57,0.85)},anchor=east,font=\small, legend columns=1, draw, fill},
     cycle list name = mycyclelist,
     mark repeat={1},
     grid=both,
     label style={font=\small},
     xtick pos=left,
     ytick pos=left,
     xmin=0, xmax=1, xtick={0,0.1,0.2,0.3,0.4,0.5,0.6,0.7,0.8,0.9,1},
     xlabel= Normalised path loss of MTDs,
     ylabel= Avg. total energy (mW),
     ylabel style={yshift=-0.5ex},
     ymin=175, ymax=375, ytick={175,200,225,250,275,300,325,350,375},
     ticklabel style={
        /pgf/number format/fixed,
        /pgf/number format/precision=5
     }
     ]
        \addplot table[x expr={\thisrow{x}/9-10/9+1}, y expr={\thisrow{y}*1000}] {avg_E_tot_b.txt};
        \addplot table[x expr={\thisrow{x}/9-10/9+1}, y expr={\thisrow{y}*1000}] {avg_E_tot_p.txt};
    \legend{Benchmark scheme, Proposed scheme}
    \end{axis}
        \end{tikzpicture}
\caption{The average total energy cost of $k$th~MTD, $\frac{1}{M}\sum_{m=1}^{M} E^{(m)}_{\text{cp+tx},k}$, per communication round versus its path loss for the benchmark and proposed schemes.}
\label{avg_E_tot}
\vspace{-0.25cm}
\end{figure}

From Fig.~\ref{avg_loss}, we observed fairness in learning despite the divergent global model quality of MTDs. Now we analyze the energy expenditure of channel-heterogeneous MTDs for the same simulation setup. In that Fig.~\ref{avg_E_tot} plots the average total energy cost of $k$th~MTD, $\frac{1}{M}\sum_{m=1}^{M} E^{(m)}_{\text{cp+tx},k}$, per communication round versus its path loss for the benchmark and proposed schemes. As expected, for the benchmark scheme the total energy cost significantly increases with the path loss severity due to the so-called near-far problem. It is because, in the vanilla FL setting, the focus is kept on the local model quality and as many as possible iterations are performed for given delay bound. In contrast, for the proposed scheme the total energy cost remains almost flat for all MTDs irrespective of the channel statistics. Thereby, as desired, the MTDs spend similar energy to learn similar quality of the global model (utility fairness). Specifically, the proposed scheme reduces the standard deviation of the energy cost across MTDs by 99\% and provides about 12.17\% reduction in the average energy cost across MTDs, as compared to the benchmark scheme. Importantly the proposed scheme achieves this utility fairness without any knowledge of the computation or transmission expenditure of MTDs at the AP or among MTDs.

\section{Conclusion}

In this paper, we have investigated the utility unfairness issue in a FL based wireless IoT network due to the device-heterogeneity.
We proposed to control the quality of the global model shared with the devices, in each round, based on their contribution and expenditure. We designed a utility function which works as a catalyst and it is used to reveal the optimal computation and transmission policies, such that the learning gain versus the cost is similar for all devices, without any knowledge of the contribution and expenditure of other devices.
Our results showed that the proposed scheme reduces the standard deviation of the energy cost across MTDs by 99\% in comparison to the benchmark scheme, while the standard deviation of the training loss of MTDs varies around 0.103. In addition, the proposed scheme provides about 12.17\% reduction in the average energy cost across MTDs.

\bibliographystyle{IEEEtran}
\bibliography{IEEEabrv,cite}

\end{document}